\documentclass[10pt,twocolumn,letterpaper]{article}

\usepackage[]{cvpr}      

\usepackage{graphicx}
\usepackage{multirow}
\usepackage{amsmath}
\usepackage{amssymb}
\usepackage{booktabs}
\usepackage{tabularx}
\usepackage{amssymb}
\usepackage{pifont}
\usepackage{siunitx}

\usepackage[pagebackref,breaklinks,colorlinks]{hyperref}
\usepackage{array}
\usepackage{booktabs}
\usepackage{tabularx}
\usepackage{xcolor}
\usepackage{listings}
\usepackage{float}
\usepackage[capitalize]{cleveref}
\crefname{section}{Sec.}{Secs.}
\Crefname{section}{Section}{Sections}
\Crefname{table}{Table}{Tables}
\crefname{table}{Tab.}{Tabs.}

\def\cvprPaperID{*****} 
\def\confName{CVPR}
\def\confYear{2022}

\begin{document}

\title{Grounded Gesture Generation: Language, Motion, and Space}

\author{
Anna Deichler$^{1}$ \quad
Jim O'Regan$^{1}$ \quad
Téo Guichoux$^{2}$ \quad
David Johansson$^{1}$ \quad
Jonas Beskow$^{1}$ \\
$^{1}$KTH Royal Institute of Technology \\
$^{2}$Sorbonne University
}

\twocolumn[{
  \maketitle
  
  \begin{center}
    \includegraphics[width=\linewidth]{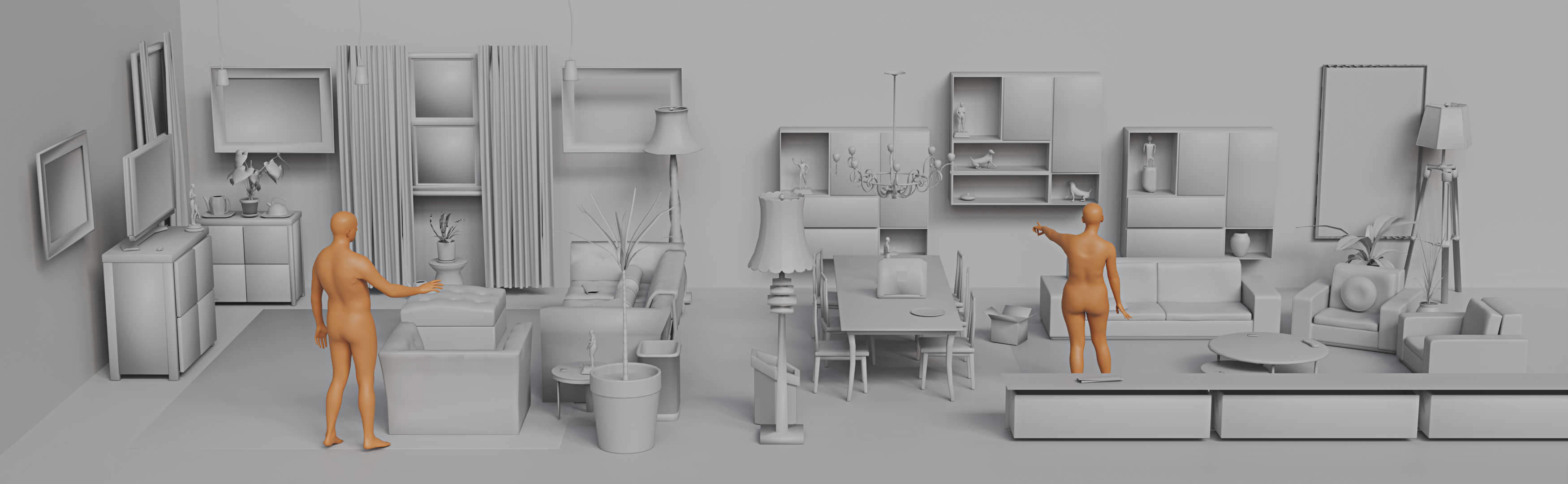}
    \label{fig:teaser}
  \end{center}
}]

\begin{abstract}Human motion generation has advanced rapidly in recent years, yet the critical problem of creating spatially grounded, context-aware gestures has been largely overlooked. Existing models typically specialize either in descriptive motion generation, such as locomotion and object interaction, or in isolated co-speech gesture synthesis aligned with utterance semantics. However, both lines of work often treat motion and environmental grounding separately, limiting advances toward embodied, communicative agents. To address this gap, our work introduces a multimodal dataset and framework for grounded gesture generation, combining two key resources: (1) a synthetic dataset of spatially grounded referential gestures, and (2) MM-Conv, a VR-based dataset capturing two-party dialogues. Together, they provide over 7.7 hours of synchronized motion, speech, and 3D scene information, standardized in the HumanML3D format. Our framework further connects to a physics-based simulator, enabling synthetic data generation and situated evaluation. By bridging gesture modeling and spatial grounding, our contribution establishes a foundation for advancing research in situated gesture generation and grounded multimodal interaction. Project page: \textit{\textcolor{magenta}{\url{https://groundedgestures.github.io/}}}. \end{abstract}

\section{Introduction}
Situated referential communication, where speech, gesture, and gaze are coordinated to ground meaning in the surrounding environment, is a core aspect of human interaction. It is essential not only for resolving ambiguities and clearly expressing communicative intent in everyday conversation, but also for pedagogical instruction, language development, and overcoming language barriers. In recent years, there has been a surge in technologies related to the generation of humanoid agents with believable and communicative behaviors. 
However, the generation of \emph{spatially grounded, context-aware gestures} for these agents remains a significant challenge.

To date, research on motion generation using deep generative models has largely focused on two separate problem domains. One is co-speech gesture generation, which aims to generate believable and semantically plausible gestures that match a verbal message in speech and/or text. These models are trained on large corpora of motion capture or video, conditioned with speech and/or text with speech, typically with no spatial information other than that of the interlocutor (if present). The other class of models are so-called ``text-to-motion'' models, where natural language prompts are used to drive human behavior. These models are trained on large and diverse datasets of human motion, and can also in some cases incorporate semantic and geometric information about the scene, allowing instructions like ``walk over to the sofa and sit down'', but these models have no concept of speech and gesture. In order to build humanoid agents that can function in real life scenarios, we need models that can integrate spatial information and communicative behavior, including speech and gesture. Progress is constrained by the lack of standardized datasets that combine motion, language, gaze, and 3D scene information in a way that supports both training and evaluation of situated behaviors. Existing datasets often treat gesture generation and environmental grounding separately, limiting advances in embodied, interactive AI systems.

One issue that hinders efforts trying to bridge these domains is discrepancy in terms of motion representation. For text-to-motion models, HumanML3D or SMPL-X are the dominant formats. In co-speech gesture, the tradition is to represent motion as skeletal joint angles (e.g. BVH-format). Recently \cite{liu2024emage} introduced SMPL-X representation for one co-speech gesture dataset, but most of the available co-speech data is still in BVH format which hinders integration tasks.

In this work, we address this gap by introducing a multimodal dataset and framework for situated referential gesture research. We combine two complementary resources, a synthetic dataset of spatially grounded referential gestures and MM-Conv, a VR-based dataset of spontaneous two-party dialogues, converted into HumanML3D-compatible format. Beyond dataset integration, we connect the data to a physics-based simulator, enabling synthetic data generation, situated evaluation, and future extensions toward embodied agents. As a proof-of-concept, we also fine-tune a motion generation model (OmniControl \cite{xie2024omnicontrol}) on the data to produce spatially controllable pointing gestures. 

Our main contributions are: \begin{itemize} \item \textbf{Standardization:} Integration of synthetic and real referential gesture datasets into the HumanML3D format, aligning with emerging standards in human-scene modeling. \item \textbf{Framework:} A proposed modular architecture that integrates spatial cues into gesture generation, supporting both synthetic data generation and situated evaluation in dynamic 3D environments. \item \textbf{Resources:} Release of over 7.7 hours of synchronized motion, speech, and scene data, supporting the study of grounded communication in immersive settings. \end{itemize}
\section{Related Work}

Research at the intersection of gesture, language, and 3D scene understanding has advanced significantly in recent years, but grounded referential gesture generation remains underexplored. Most progress has occurred in two relatively separate domains: (1) language-conditioned motion generation, which focuses on producing motion from textual instructions, and (2) co-speech gesture generation, which aims to synchronize gestures with speech for naturalistic communication. In this section, we review both domains and highlight their limitations for building communicative, scene-aware humanoid agents.   

 A structured comparison of co-speech and motion-language datasets, including modality coverage, representation format, and communicative focus, is provided in Table \ref{tab:motiondatasets}.

\begin{table*}[t]
\footnotesize
\centering
\renewcommand{\arraystretch}{1.2}
\caption{Comparison of human motion and gesture datasets highlighting modalities, text types, settings, and scales.}
\begin{tabularx}{\textwidth}{@{}>{\raggedright\arraybackslash}p{2.9cm} >{\raggedright\arraybackslash}p{3.2cm} >{\raggedright\arraybackslash}p{1.8cm} >{\raggedright\arraybackslash}p{1.9cm} >{\raggedright\arraybackslash}p{1.5cm} >{\raggedright\arraybackslash}p{4.4cm}@{}}
\toprule
\textbf{Dataset} & \textbf{Modalities} & \textbf{Text Type} & \textbf{Motion Representation} & \textbf{Duration} & \textbf{Focus} \\
\midrule
Grounded Gestures (Ours) & Motion, Speech, Scene Info & Referential utterances & SMPL-X & 7.7 h & Referential/ conversational gestures \\
MM-Conv \cite{deichler2024mm} & Motion, Speech, Gaze, Scene Graphs & Spontaneous dialogue & BVH & 6.5 h & Spontaneous referential communication \\
HumanML3D \cite{guo2022humanml3d} & Text, 3D motion & Descriptive captions & SMPL+H/DMPL & 28.59h & Text-conditioned general motion \\
HUMANISE \cite{wang2022humanise} & Text, 3D motion & Descriptive captions & SMPL+H/DMPL & 28.59h & Text-conditioned general motion \\

BEAT \cite{liu2022beat} & Text, Speech, Motion & Transcripts & BVH & 76h & Co-speech beat and iconic gestures \\
BEAT2 \cite{liu2024emage} & Text, Speech, Motion & Transcripts & SMPL-X & 60h & Co-speech beat and iconic gestures \\
TalkSHOW \cite{yi2023generating} & Text, Speech, Motion & Transcripts & SMPL-X & 27h & Video extracted Co-speech gestures \\
Trinity Speech Gesture (I \cite{IVA:2018} and II\cite{ginosar2019learning}) & Text, Speech, Motion & Transcripts & BVH &  6h + 4h & Co-speech gestures \\
KIT ML \cite{plappert2016kit} & Text, MoCap & Action labels & SMPL\cite{guo2022generating} &  11.23h & Action-mapped motions \\
\bottomrule
\end{tabularx}
\label{tab:motiondatasets}
\end{table*}

\subsection{Language-conditioned human motion generation}

The AMASS (Archive of Motion Capture as Surface Shapes) \cite{mahmood2019amass} dataset is a comprehensive collection of human motion data, integrating 15 different optical marker-based motion capture datasets into a unified framework. It provides realistic 3D human meshes represented by a rigged body model, making it valuable for various applications in animation, visualization, and deep learning. It contains over 11k motion sequences, 40 hours of data, more than 300 individuals.

HumanML3D \cite{guo2022generating} is derived from AMASS and HumanAct12 dataset and is designed to bridge natural language and 3D human motions, facilitating tasks like text-to-motion generation and motion captioning. It contains over 14k motion sequences, approximately 28.6 hours of data. The average motion length is 7.1 seconds, while average description length is 12 words. The minimum and maximum duration are 2s and 10s respectively. In terms of the textual descriptions, their average and median lengths are 12 and 10, respectively. 

 HUMANISE \cite{wang2022humanise}  introduces a large-scale synthetic dataset for language-conditioned human motion generation in 3D scenes. It aligns 19,600 AMASS motion sequences with richly annotated ScanNet 3D scans and auto-generated language descriptions that specify both actions and interacting objects (e.g., “sit on the armchair near the desk”). Spanning 643 indoor scenes, the dataset enables research on spatial reasoning between motion, object geometry, and language for embodied AI.

Recent advances in human motion generation increasingly rely on diffusion models, which have demonstrated strong capabilities in producing diverse, high-quality motion sequences. Language-conditioned models such as those used in HUMANISE, as well as scene-aware and language-guided frameworks like TeSMo \cite{yi2024generating}, now commonly adopt diffusion-based architectures. This reflects the growing dominance of diffusion models for generating realistic and contextually grounded human motions. 

\subsection{Co-Speech Gesture Generation}
Co-speech gesture generation focuses on synthesizing naturalistic hand and body movements that accompany spoken language. Early approaches often relied on rules or heuristics in combination with procedural animation, and could support different types of co-speech gestures, including pointing to specific spatial locations \cite{kappagantula2019automatic,lester1999deictic,noma2002design,rickel1999animated}. The motion produced by these systems often had an unmistakable synthetic quality. With the availability of large-scale datasets of synchronized speech and motion data, data-driven methods have become the dominant paradigm. These datasets fostered the development of models that achieve high naturalness and improved speech-gesture synchronization, however, at the cost of lacking control and spatial awareness.

The Trinity Speech-Gesture Dataset \cite{ferstl2018investigating} contains 23 recordings totaling 244 minutes of synchronized full-body motion capture and audio of a male native English speaker delivering spontaneous monologues on various topics. The Talking with Hands16.2M dataset \cite{lee2019talking} provides approximately 50 hours of synchronized full-body and detailed hand motion, recorded during natural dyadic conversations in a motion capture laboratory setting. The BEAT dataset \cite{liu2022beat} offers large-scale multimodal data combining speech, text, and full-body motion. The BEAT2 \cite{liu2024emage} dataset combines SMPL-X body estimated using MoSh \cite{loper2014mosh} with head parameters estimated using FLAME \cite{li2017flame}.

Diffusion models have also shown strong performance in gesture generation. For example, GestureDiffuClip \cite{ao2023gesturediffuclip} related approaches use diffusion processes to model complex motion distributions conditioned on speech (audio and text) embeddings. Other works have incorporated contrastive learning strategies to better align motion and speech representations \cite{deichler2023diffusion}.

While these methods have led to substantial improvements in the realism and timing of gestures relative to speech, most focus on isolated gesture generation without grounding gestures in a shared spatial or interactional context. As a result, they often neglect the spatial communicative functions of gestures essential for situated communication.

\section{Dataset}
We aim to standardize situated gesture generation datasets. To this end, we build on two separate motion-capture-based datasets, which contain conversational and pointing gestures. Together, these datasets enable the development of models that combine co-speech and deictic gestures in situated settings.






\begin{itemize}
    \item The \textbf{synthetic pointing gesture dataset}  consists of isolated pointing gestures, with accurate 3D target locations captured via marker-based motion capture. 
    \item The \textbf{VR-based referential dataset (MM-Conv)}  contains natural conversations recorded in virtual reality environments, using motion capture, where participants interact with shared virtual spaces. 
\end{itemize}

\subsection{Motion data preparation}
\label{sec:motion_prep}
We use \cite{xu2024inter} to transform the global joint position data from the motion capture system into SMPL-X \cite{SMPL-X:2019} format. Then we follow the steps described in \cite{guo2022generating} to transform the SMPL-X data format into the HumanML3D format.

In HumanML3D, each body pose is parameterized by a vector 
\(\mathbf{x}\in\mathbb{R}^{263}\), whose entries are
\[
\underbrace{\omega_r}_{1}
\;,\;
\underbrace{\mathbf{v}_r}_{2}
\;,\;
\underbrace{y_r}_{1}
\;,\;
\underbrace{p}_{21\times3}
\;,\;
\underbrace{R}_{21\times6}
\;,\;
\underbrace{\dot R}_{22\times3}
\;,\;
\underbrace{c}_{4}
\]
corresponding respectively to root rotational velocity, root linear velocity, root vertical position, rotation-invariant joint positions, joint rotation parameters, joint rotational velocities, and binary foot-contact indicators, with total dimensionality
\(\dim(\mathbf{x})=1+2+1+21\cdot3+21\cdot6+22\cdot3+4=263\).
Pelvic translation is encoded as the displacement from the preceding frame, while all other joints are expressed relative to the pelvis; this representation facilitates learning and leads to more realistic synthesized motions \cite{xie2024omnicontrol}. 
The motion sequences are sampled at 20 frames per second.

\subsection{Synthetic pointing gesture data generation}
The paper \cite{deichler2024incorporating}  proposes integrating spatial context into virtual agents' gesture generation by combining co-speech and pointing gestures. A synthetic dataset was created to extend current speech-driven models with spatially grounded behaviors, enabling agents to perform deictic gestures aligned with speech.  We follow the main generation steps outlined in the paper to create semantically aligned data with the VR-based referential dataset \cite{deichler2024mm}:
\begin{enumerate}
    \item Take the augmented pointing dataset from \cite{deichler2022towards}.
    \item Generate new demonstrative utterances using a VLM.
    \item Synthesize utterances with TTS.
    \item Temporally align the synthesized utterances with the pointing motions using the Hungarian algorithm, including padding to match durations.
\end{enumerate}

\begin{figure}[t]
  \centering
  \includegraphics[width=0.8\linewidth]{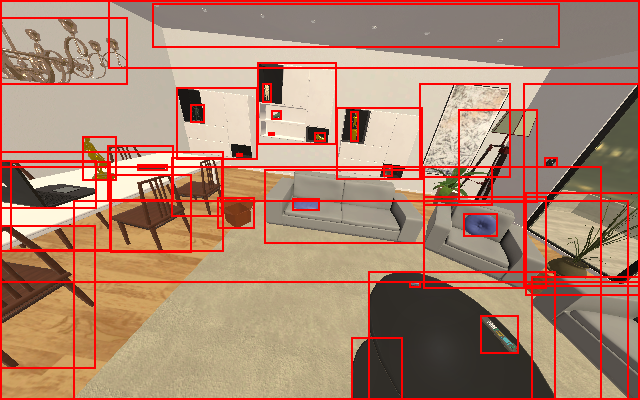}
  \caption{Ground truth bounding boxes of objects in current view, extracted from simulator\cite{kolve2017ai2}.}
  \label{fig:mask_with_bboxes}
\end{figure}

We  transform the pointing dataset BVH files using the process described in section~\ref{sec:motion_prep}. This results in 1135 motion files with average duration of 4.85 seconds.

We use GroundingGPT \cite{li2024groundinggpt} to adjust the prompt generation for the MM-Conv dataset. Specifically, we rendered an RGB image from the character's viewpoint and extracted the ground truth segmentation mask with object labels from the simulator (Figure \ref{fig:mask_with_bboxes}). We randomly selected one object and passed it to the VLM using the prompt in Appendix~\ref{appendix:exophoric_prompt}. We generate exophoric referring expressions involving demonstratives ('this' or 'that') in a conversational style. Examples of generated outputs:
\begin{itemize}
    \item \textit{See that chair over there, it's the perfect spot for a relaxing read.}
    \item \textit{Look at that painting on the wall over there, it's really beautiful.}
\end{itemize}

\begin{figure}[t]
  \centering
  \includegraphics[width=0.9\linewidth]{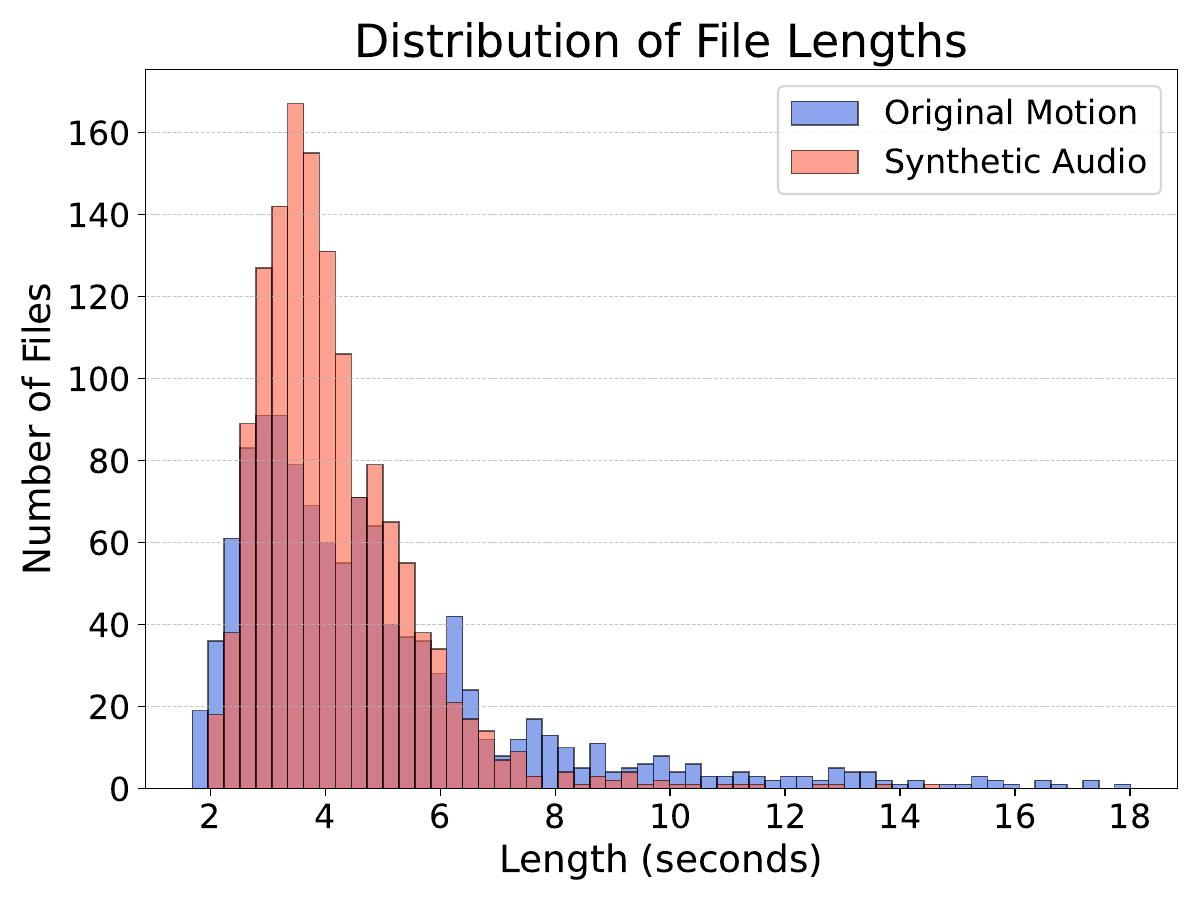}
  \caption{Distribution of file lengths for synthetic audio and original motion segments.}
  \label{fig:histogram_lengths}
\end{figure}

We used \cite{chen2024f5} to generate natural-sounding synthetic speech from the generated sentences, using voice cloning of the original actors. The resulting speech samples are then processed using WhisperX \cite{bain2022whisperx} to obtain word-level timestamps. The output of WhisperX is additionally compared to the original text using a Word Error Rate threshold of 0.3, to filter for major divergences in generation, discarding any samples that deviate too much from the original sentence. 94 samples were removed by this step. The remaining 1406 files were matched with the 1135 pointing gesture motions using the Hungarian algorithm based on the demonstrative location. Figure~\ref{fig:histogram_lengths} shows the distribution of durations for the synthesized audio clips and the original pointing gesture segments.

Examples of aligned, synthesized pointing gestures are provided in the supplementary material. Additional illustrative examples can be found in Appendix~\ref{appendix:synth_ref_images}.

\begin{figure*}[t]
  \centering
  \includegraphics[width=0.6\textwidth]{
  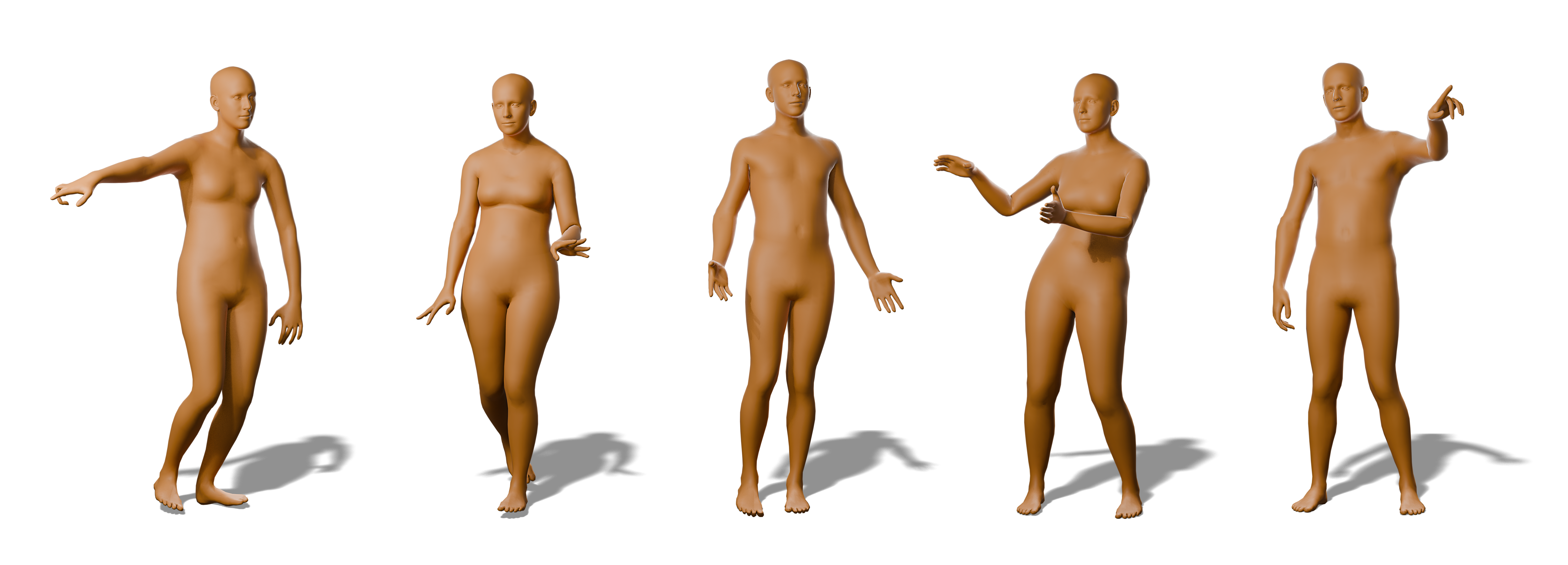}
  \caption{Rendered SMPL-X frames from the VR-based referential dataset (MM-Conv).}
  \label{fig:referential_poses}
\end{figure*}

\subsection{VR-based referential dataset}
The MM-Conv dataset \cite{deichler2024mm} is a multimodal conversation dataset designed for training speech-driven gesture generation models. It contains motion capture recordings of human conversations conducted in virtual reality (VR) environments using the Ai2-Thor simulator \cite{kolve2017ai2}. Participants interact naturally, referring to shared virtual environments that include objects and spatial landmarks. The dataset includes synchronized speech audio, text transcripts, and full-body motion capture data encompassing hands, head, and body movements. MM-Conv is specifically designed to link language, motion and space, providing a rich foundation for research in co-speech gesture generation and situated communication tasks.

\paragraph{Reference Annotation Output.} 

The output of the reference annotation process from \cite{multimodalreferencing2025} is stored in a structured JSON format. 
Each utterance is associated with a snippet of transcribed speech and a list of extracted references. 
For each reference, we store the resolved object ID, the original phrase, plurality information (e.g., ``single'' or ``multiple''), the reference type (e.g., ``exact'' noun phrase), and the phrase's start and end timestamps (in seconds) within the utterance. 
An example is shown on Listing~\ref{lst:json-ref-annotation}.
\begin{lstlisting}[float, floatplacement=H, caption={Example of a structured JSON annotation for a referential utterance.}, label={lst:json-ref-annotation}]
"9": {
  "snippet": "Oh yeah, yeah, I stole this from the Louvre.",
  "references": [
    {
      "ref_id": 0,
      "phrase": "this",
      "resolved_ref": "Painting_eda6d5e1",
      "plurality": "single",
      "original_label": "pronominal",
      "object_id": "Painting_eda6d5e1",
      "phrase_start": 38.86,
      "phrase_end": 39.04
    }
  ],
  "start": 37.880602861595214,
  "end": 39.702,
  "high_level": {
    "current_topic": "Painting_eda6d5e1",
    "topic_duration_id": 1
  }
}
\end{lstlisting}



Each utterance also includes high-level topic annotations, which track the object of interest across multiple utterances. This structure enables fine-grained multimodal grounding, supporting tasks such as reference resolution, object localization, and dialog analysis.

\paragraph{Motion Segmentation Output.}

The motion data associated with each recording is segmented into \textit{referential} and \textit{non-referential} motion segments. Referential segments correspond to utterances that contain at least one resolved reference to an object in the environment. Each referential segment is extracted based on the start and end timestamps of the utterance, covering the motion that co-occurs with the referential expression. Non-referential segments capture all remaining motion, including both speech without object references and silent periods between utterances. To ensure complete coverage, non-referential motion is computed as the complement of the referential intervals over the full recording timeline. This partitioning enables training and evaluation of models for both grounded, reference-driven gesture generation and background or spontaneous motion modeling.

Following the filtering practices in prior motion-language datasets such as HumanML3D~\cite{guo2022generating}, we apply duration-based filtering to ensure data consistency. Referential motion segments shorter than 0.5 seconds or longer than 20 seconds are discarded. Non-referential motion is segmented into clips strictly between 0.5 and 20 seconds by splitting longer intervals and discarding very short ones. This process improves dataset uniformity and facilitates stable training for generative models.

Table~\ref{tab:dataset_stats_filtered} summarizes the motion segment statistics after filtering, with 2,394 referential and 2,721 non-referential motion clips. The overall dataset spans 6.14 hours, with a mean clip duration of 4.32 seconds across 64 joints in 3D space. 

Table~\ref{tab:reference_stats_filtered} summarizes the distribution of references across referential utterances. Out of 2,394 referential utterances, the majority involve a single reference, while a substantial proportion involve multiple references, highlighting the presence of both simple and complex referential structures. On average, each utterance includes 1.65 annotated references, enabling modeling of both simple and complex referential structures.  At the reference level, most annotations point to single objects, with a smaller but notable fraction referring to multiple objects. These statistics provide insight into the diversity and richness of referential behavior in the dataset.

\begin{table}[h]
\centering
\footnotesize
\begin{tabular}{lcc}
\toprule
Statistic & REF & NON-REF \\
\midrule
Number of clips & 2394 & 2721 \\
Total duration (min) & 169.41 & 198.75 \\
Mean duration (sec) & 4.25 & 4.38 \\
Std duration (sec) & 3.56 & 3.65 \\
Median duration (sec) & 3.18 & 4.50 \\
Motion dimensionality & \multicolumn{2}{c}{64 joints $\times$ 3D} \\
\midrule
Total dataset duration (min) & \multicolumn{2}{c}{368.16 (6.14 hours)} \\
Overall mean duration (sec) & \multicolumn{2}{c}{4.32} \\
\bottomrule
\end{tabular}
\caption{Statistics of the dataset after filtering, broken down into referential and non-referential motion segments.}
\label{tab:dataset_stats_filtered}
\end{table}

\begin{table}[h]
\centering
\footnotesize
\begin{tabular}{lc}
\toprule
Statistic & Value \\
\midrule
Total referential utterances & 2394 \\
Utterances with 1 reference & 1377 (57.5\%) \\
Utterances with $>$1 reference & 1017 (42.5\%) \\
Total references & 3955 \\
References (single object) & 2991 (75.6\%) \\
References (multiple object) & 964 (24.4\%) \\
Average references per referential utterance & 1.65 \\
\bottomrule
\end{tabular}
\caption{Statistics of references across referential utterances after filtering.}
\label{tab:reference_stats_filtered}
\end{table}

\begin{figure}[t]
  \centering
  \includegraphics[width=0.8\linewidth]{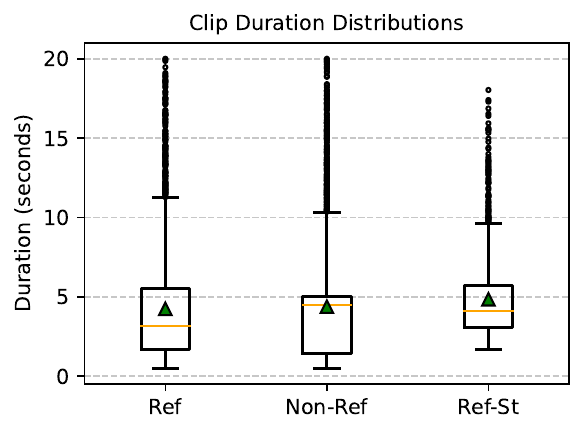}
  \caption{Distribution of motion clip durations for referential and non-referential segments after filtering. 
  The boxplots show median (orange line), mean (green triangle), and outliers (black dots).}
  \label{fig:clip_duration_boxplot}
\end{figure}



\subsection{Dataset Summary}

Overall, the combined corpus spans over 7.7 hours of 3D motion data, with consistent clip durations and detailed reference labels. These resources enable the development and evaluation of generative models for co-speech and deictic gesture generation in situated contexts. Figure~\ref{fig:clip_duration_boxplot} illustrates the distribution of motion clip durations across referential and non-referential segments, summarizing the temporal structure of the dataset after filtering.

\subsection{Training Splits}
\label{sec:splits}
To facilitate model training under different scenarios, we define three training splits with increasing data coverage:
\begin{itemize}
    \item \textbf{ST only}: Only synthetic pointing motion sequences.
    \item \textbf{REF + ST}: Referential motion segments combined with synthetic pointing data.
    \item \textbf{ALL (REF + ST + NONREF)}: Full dataset including referential, synthetic pointing, and non-referential motion clips.
\end{itemize}

Table~\ref{tab:training_splits} summarizes the statistics of the three splits, highlighting the number of motion clips, total duration, and duration distribution characteristics.

\begin{table}[h]
\centering
\footnotesize
\begin{tabular}{lccc}
\toprule
Statistic & ST & REF+ST & ALL \\
\midrule
\# clips & 1135 & 3529 & 6250 \\
Total (min) & 91.70 & 261.11 & 459.86 \\
Mean (sec) & 4.85 $\pm$ 2.65 & 4.45 $\pm$ 3.10 & 4.35 $\pm$ 3.29 \\
\midrule
Motion dim & \multicolumn{3}{c}{64 joints $\times$ 3D} \\
\midrule
Total (min) & \multicolumn{3}{c}{459.86 (7.66 hours)} \\
Mean  (sec) & \multicolumn{3}{c}{4.45} \\
\bottomrule
\end{tabular}
\caption{Statistics of the three defined training splits: ST only (synthetic pointing), REF + ST (referential and synthetic pointing), and ALL (referential, synthetic pointing, and non-referential).}
\label{tab:training_splits}
\end{table}

\begin{figure}[t]
  \centering
  \begin{subfigure}{0.48\linewidth}
    \includegraphics[width=\linewidth]{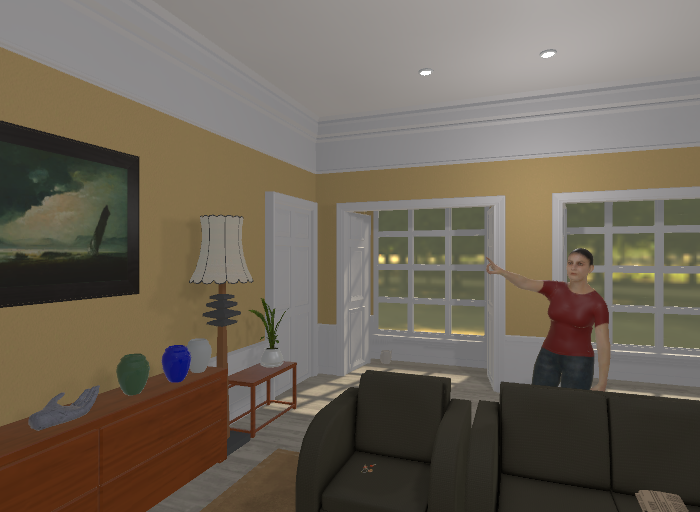}
    \caption{Rendered RGB 3rd person view.}
    \label{fig:example_render_3rd}
  \end{subfigure}
  \hfill
  \begin{subfigure}{0.48\linewidth}
    \includegraphics[width=\linewidth]{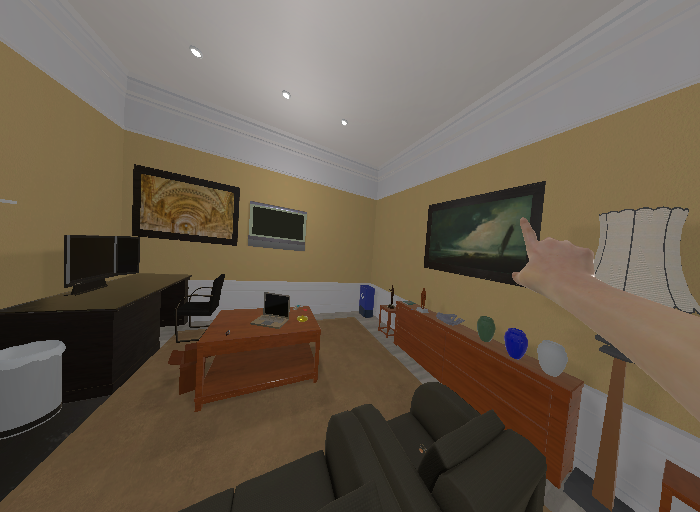}
    \caption{Rendered RGB 1st person view.}
    \label{fig:color_image}
  \end{subfigure}
  \caption{Examples of rendered data sample of SMPL-X character in simulator \cite{kolve2017ai2}.}
  \label{fig:example_render_1st}
\end{figure}

    \section{Experiments}

    In situated conversational agents, the generation task shifts from goal-driven locomotion to understanding spatial references in dialog using a vision-language model and generating referential language combined with gestures. An open question is the appropriate spatial semantic representation for gesture generation: does the gesture module require the same spatial granularity as humanoid locomotion, or a more semantically nuanced representation? This affects how gestures align with communicative intent in 3D scenes.
        
\begin{figure}[h]
      \centering
      \includegraphics[width=0.9\linewidth]{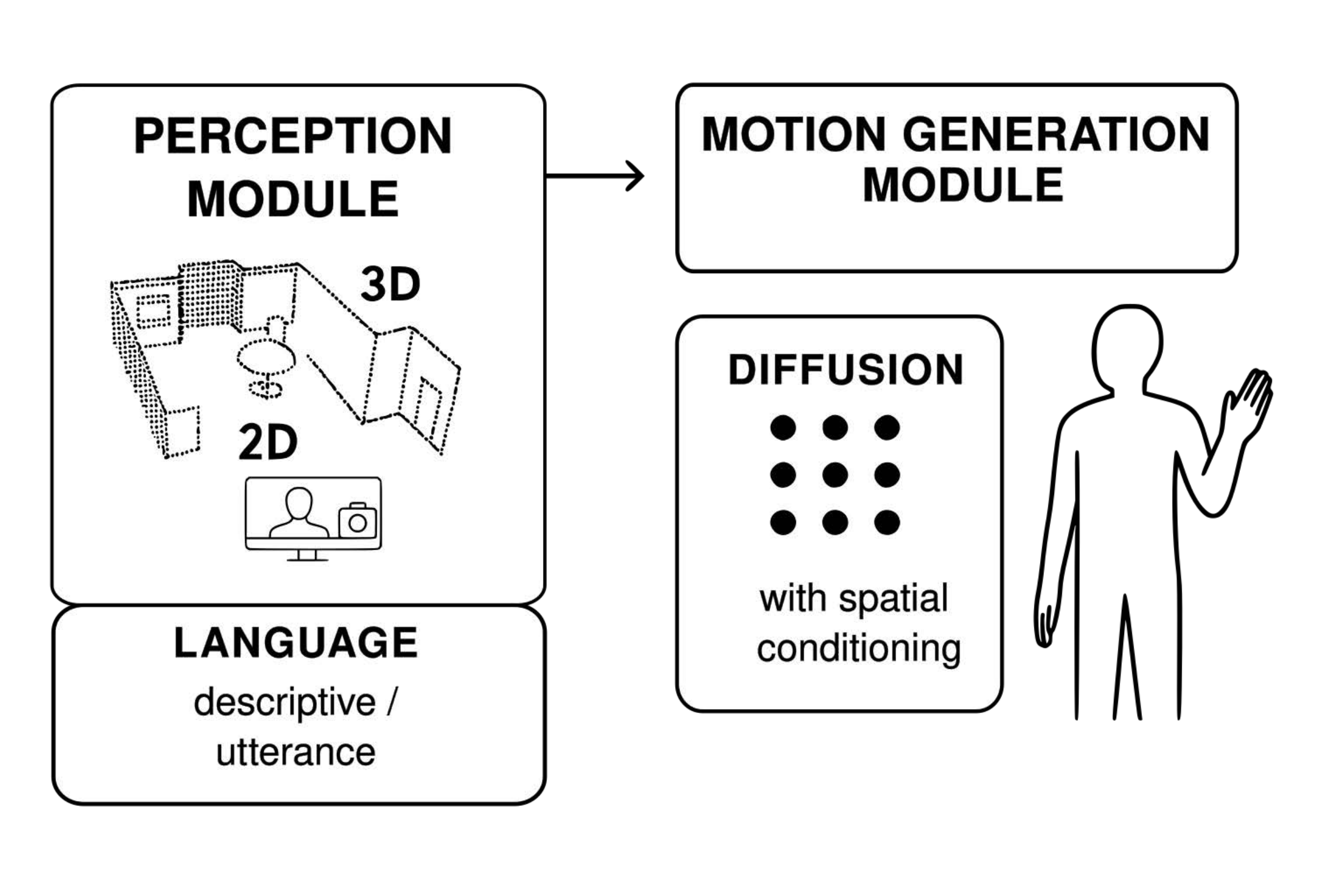}
      \caption{Proposed framework.}
      \label{fig:framework}
    \end{figure}

    Recent advances in controlled motion generation suggest that modular systems, where perception and generation are decoupled, offer better scalability and interpretability \cite{wang2024move,li2024controllable}. Inspired by these approaches, we propose a modular framework where the system first derives referenced objects from the scene (e.g., via bounding boxes or graphs) and predicts a spatial cue, such as a referent’s location, using a VLM or 3D scene understanding module (Figure \ref{fig:framework}). Conditioned on the detected spatial cue, a generative motion model then produces a gesture that grounds the intent toward the detected location. This decomposition enables flexible training, better interpretability, and supports extensions toward fully embodied agents in complex 3D environments.
    
    \subsection{Motion generation with spatial conditioning}
    
    OmniControl \cite{xie2024omnicontrol} is a diffusion-based motion generation model trained on the HumanML3D dataset, which enables precise control over any joint in time, guided by text prompts and spatial constraints. It uses joint-conditioned guidance during training to align generated motions with the specified constraints while maintaining naturalness in the generated motion. This approach allows for flexible and high-fidelity motion synthesis, outperforming previous methods limited to pelvis control.
    
    To incorporate spatial control, OmniControl introduces a spatial guidance mechanism that enforces joint-level constraints during the denoising process. At each diffusion timestep \( t \), given a partially noised motion sequence \( \mathbf{x}_t \), the model predicts the clean motion \( \hat{\mathbf{x}}_0 \) by minimizing a composite loss function:
    \begin{equation}
    \mathcal{L}_{\text{total}} = \mathcal{L}_{\text{denoise}} + \lambda_{\text{spatial}} \mathcal{L}_{\text{spatial}} + \lambda_{\text{realism}} \mathcal{L}_{\text{realism}},
    \end{equation}
    where \( \mathcal{L}_{\text{denoise}} \) is the standard denoising loss over the motion sequence, 
    \( \mathcal{L}_{\text{spatial}} = \| \mathbf{x}_t^j - \mathbf{g}_t^j \|_2 \) 
    is the spatial loss ensuring that the generated joint positions \( \mathbf{x}_t^j \) match the target positions \( \mathbf{g}_t^j \) for controlled joints \( j \), and \( \mathcal{L}_{\text{realism}} \) encourages plausible and coherent motion trajectories. The spatial guidance is implemented by converting generated motions to global coordinates and aligning them with the spatial targets, allowing control over arbitrary joints and frames without retraining. This is especially relevant for grounded gesture generation, since the target representation is naturally defined in the global coordinate frame, allowing the model to produce gestures that are spatially aligned with the referent’s position in the shared environment.
    


    \subsection{Training Configurations}
    
    We fine-tuned OmniControl \cite{xie2024omnicontrol} on three subsets of our dataset, defined in section \ref{sec:splits}, and evaluated models using three control joints —pelvis, left wrist, right wrist— chosen for their importance in modeling pointing gestures. Each configuration was evaluated in two modes: the base model trained exclusively on HumanML3D, and its fine-tuned counterpart adapted to each of the three dataset splits.

    \subsection{Results}

    Table~\ref{tab:fid_control} summarizes the Fréchet Inception Distance (FID) and Control $L_2$ error for each experimental configuration, using the same evaluation protocol as Guo et al. \cite{guo2022humanml3d}. The FID measures the naturalness of the generated motion by quantifying the discrepancy between the distributions of generated versus ground-truth motions in the Inception latent space, while the Control $L_2$ error measures the average Euclidean distance between the model’s predicted joint positions and the provided control hints. Lower values indicate better motion quality and control accuracy. Fine-tuning consistently improves both metrics across all training subsets and joints, demonstrating the benefit of adapting the pretrained OmniControl model to task-specific motion distributions.

\begin{table}[t]
\centering
\footnotesize 
\setlength{\tabcolsep}{4pt} 
\caption{FID and Control $L_2$ error (↓) for base model (OmniControl) vs.\ fine-tuned models across the training sets.}
\label{tab:fid_control}
\begin{tabular}{
  l
  l
  S[table-format=1.2]
  S[table-format=1.3]
  S[table-format=1.2]
  S[table-format=1.3]
}
\toprule
\multirow{2}{*}{Eval. Set} & \multirow{2}{*}{Controlled Joint} & \multicolumn{2}{c}{Base} & \multicolumn{2}{c}{Fine-tuned} \\
\cmidrule(lr){3-4} \cmidrule(lr){5-6}
  & & {FID ↓} & {$L_2$ ↓} & {FID ↓} & {$L_2$ ↓} \\ 
\midrule
\multirow{3}{*}{ST only} 
    & Pelvis      & 6.37 & 0.036 & 1.72 & 0.020 \\
    & Left Wrist  & 3.82 & 0.114 & 0.82 & 0.060 \\
    & Right Wrist & 4.60 & 0.116 & 0.65 & 0.058 \\
\midrule
\multirow{3}{*}{REF + ST} 
    & Pelvis      & 5.75 & 0.112 & 1.03 & 0.034 \\
    & Left Wrist  & 3.99 & 0.099 & 1.56 & 0.064 \\
    & Right Wrist & 4.26 & 0.129 & 1.41 & 0.109 \\
\midrule
\multirow{3}{*}{ALL} 
    & Pelvis      & 7.96 & 0.125 & 2.00 & 0.055 \\
    & Left Wrist  & 4.08 & 0.113 & 1.25 & 0.099 \\
    & Right Wrist & 4.49 & 0.132 & 0.84 & 0.123 \\
\bottomrule
\end{tabular}
\end{table}
    
    Notably, the ST only fine-tuned model achieves the lowest overall errors, particularly for wrist joints (FID: 0.65--0.82, Control $L_2$: 0.058--0.060), highlighting the benefit of training on focused, synthetic pointing data. Incorporating referential data in the REF + ST split leads to improved pelvis control (FID: 1.03 vs. 1.72), suggesting that referential clips help better model full-body coordination. However, this comes with a modest increase in wrist error.
    
    The ALL split, which includes the most diverse set of motions (referential, synthetic, and non-referential), results in slightly higher FID and Control $L_2$ values for most joints. This suggests a trade-off: while broader training data improves generalization, it may introduce variance that slightly reduces precision for fine-grained control joints. Nevertheless, fine-tuned models remain consistently superior to pretrained ones across all splits, underscoring the importance of task-adaptive training.
    
    We also conduct initial evaluations on pointing gesture generation. Specifically, we select pointing motions from the test set, extract the joint positions of the wrist and elbow at the peak pointing frame, and provide these as fixed targets to the OmniControl algorithm. The resulting motion sequences, which are included in the supplementary material, demonstrate that fine-tuning on pointing gesture data contributes to generating more natural and conversational full-body motion. As a clear example, in {\footnotesize\texttt{sample24.mp4}} in the supplementary material, the base model performs a \emph{punching motion} to reach the target pose, while the fine-tuned model performs a pointing gesture. This is promising and highlights the model's capacity to adapt to spatial constraints. However, this evaluation does not fully reflect the desired behavior in pointing tasks, where the goal is not to match absolute joint positions but to produce accurate pointing directions. Achieving this requires modifying the training procedure and redefining the spatial loss in OmniControl to enforce directional alignment, e.g., by aligning the vector from the elbow to the wrist with the vector toward the referent, rather than relying solely on positional targets.
\section{Discussion}

We standardized and extended two complementary motion capture datasets to advance research in \emph{grounded gesture generation}. The first is a synthetic dataset consisting of spatially grounded pointing gestures paired with demonstrative utterances generated via vision-language prompting and aligned speech synthesis. The second is a VR-based referential dataset capturing natural, multimodal dialogue, annotated with object references and conversational structure. Both datasets are converted into a unified SMPL-X and HumanML3D-compatible format and segmented into referential and non-referential clips. The synthetic data supports controlled experimentation on isolated deictic gestures, while the VR dataset enables the study of gesture generation in dynamic, conversational settings where language, motion, and spatial context interact.

Our current evaluations primarily focus on isolated pointing gestures. While this serves as a useful baseline for spatial grounding, future work should extend to more temporally complex behaviors that better reflect the expressive range of referential communication. Nonetheless, our results demonstrate that fine-tuning on task-specific data significantly improves motion quality and control, laying a strong foundation for further advances.

To build on this foundation, future directions include:

\paragraph{Motion modeling.} While models like OmniControl can be effectively fine-tuned for referential gesture generation, there remains room for improving how spatial intent is represented. The spatial constraints and position-based losses in OmniControl need to be re-evaluated to better capture the communicative function of gestures. Additionally, pretraining on general motion corpora like AMASS may be suboptimal for communicative behaviors. Subsampling AMASS to retain only socially relevant sequences, or integrating our data with multimodal datasets like BEAT2, could support more effective representation learning for grounded interaction.

\paragraph{Multimodal conditioning.} Audio and speech conditioning are essential for co-speech gesture generation. To move toward full referential communication, future models should leverage synchronized speech input alongside spatial information. Our dataset is well-suited for such multimodal training regimes, combining natural dialogue, annotated references, and full-body motion in shared 3D environments.

\paragraph{Perception}
A critical component in this pipeline is the perception module, which provides spatial conditioning signals to the motion generator. These signals may be derived from 2D object detectors, depth maps, or 3D scene graphs, depending on the available modality. In our setup, spatial targets are extracted using a vision-language model or 3D simulation environment and projected into world coordinates. Future work could investigate the effect of different spatial representations, such as bounding boxes, keypoints, or object centroids, on gesture quality and grounding accuracy. Better perception modules may also support dynamic referent tracking and enable learning of spatial saliency over time.

{\small
\bibliographystyle{ieee_fullname}
\bibliography{egbib}
}

\section{Appendix}
\subsection{Exophoric Demonstrative Generation Prompt}
\label{appendix:exophoric_prompt}

The following prompt was used to generate exophoric referring expressions involving demonstratives (`this' or `that') in a conversational style, using GPT-4o:

\begin{lstlisting}[caption={Prompt used for generating exophoric demonstrative references using GPT-4o.}, label={lst:exophoric-generation}]
\centering 
Demonstratives in exophoric reference are used to indicate concrete physical entities in space. 
For example, I was thinking sitting down. See that bench there, we could sit there. 
Here 'that' would be the demonstrative.

I want you to generate examples like the following, which connect to a story. 
The rules are that each generated sentence should contain one exophoric reference.

Examples:
- "See that bench over THERE, we could sit there."
- "Look in the middle of the room, THAT small dog is my new best friend."
- "We could get there faster. See THAT bike over there, we could take that one."
- "Look at the top shelf. THAT is the vase I wanted to buy."

Here's a bounding box: {bbox}, which highlights the object '{chosen_obj['name']}'. 
Can you write a casual, natural-sounding sentence that mentions '{chosen_obj['name']}' 
using either 'this' or 'that'? 
For example, something like: "And I haven't told you, but that table over there? I inherited it from my mother."
\end{lstlisting}
\subsection{Examples of synthetic pointing gestures.}
\label{appendix:synth_ref_images}

\begin{figure*}[t]
  \centering

  \begin{minipage}{\textwidth}
    \centering
    \includegraphics[width=0.99\textwidth]{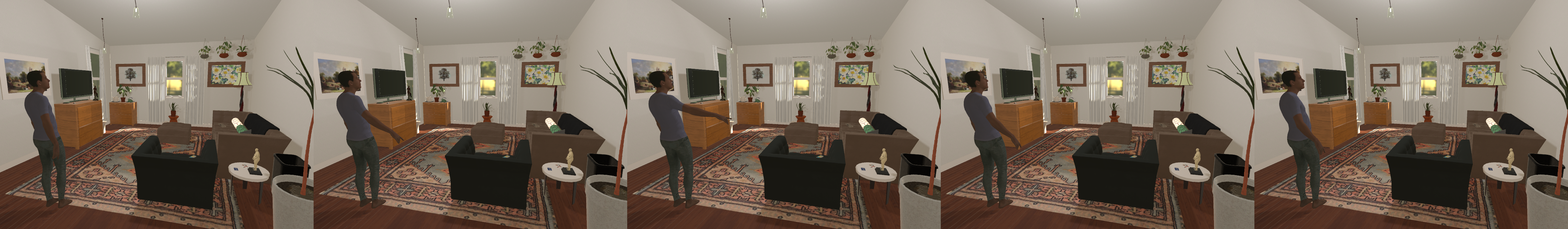}
    \subcaption{Example one of rendered third person view of an aligned synthetic pointing example.}
    \label{fig:sub1}
  \end{minipage}
  
  \vspace{0.3em}
  
  \begin{minipage}{\textwidth}
    \centering
    \includegraphics[width=0.99\textwidth]{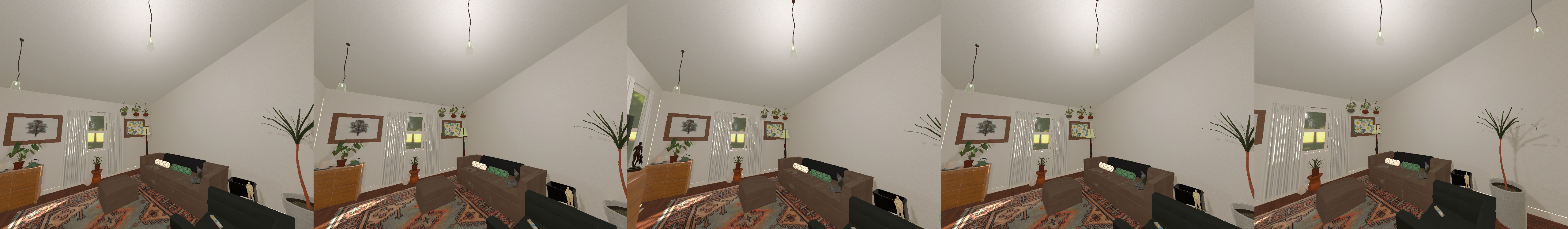}
    \subcaption{Example one of rendered first person view of an aligned synthetic pointing example.}
    \label{fig:sub2}
  \end{minipage}

  \vspace{0.3em}
  
  \begin{minipage}{\textwidth}
    \centering
    \includegraphics[width=0.99\textwidth]{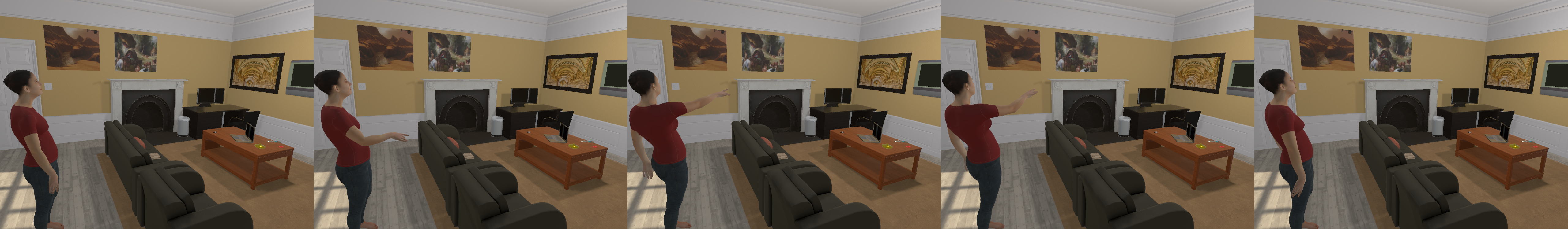}
    \subcaption{Example two of rendered third person view of an aligned synthetic pointing example.}
    \label{fig:sub3}
  \end{minipage}

  \vspace{0.3em}

  \begin{minipage}{\textwidth}
    \centering
    \includegraphics[width=0.99\textwidth]{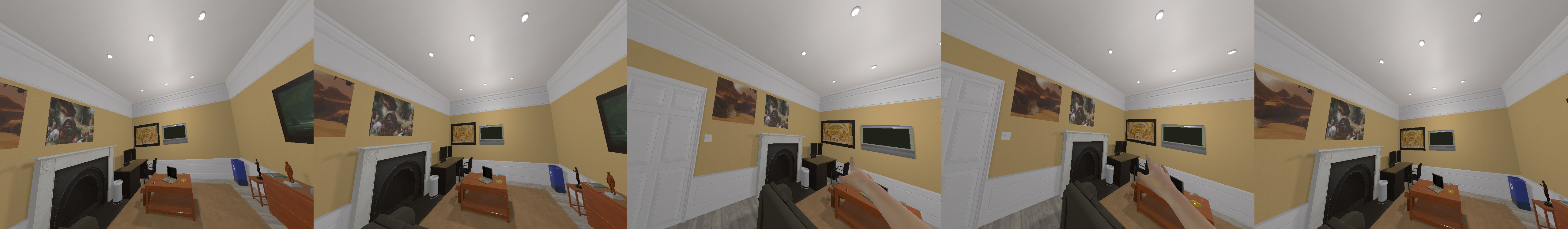}
    \subcaption{Example two of rendered first person view of an aligned synthetic pointing example.}
    \label{fig:sub4}
  \end{minipage}

  \caption{Rendered aligned synthetic pointing gesture examples in third and first person views.}
  \label{fig:synth_referential_poses}
\end{figure*}

\end{document}